\documentclass[runningheads]{llncs}

\usepackage{eccv}

\usepackage{eccvabbrv}

\usepackage{graphicx}
\usepackage{booktabs}

\usepackage[accsupp]{axessibility}  %

\usepackage[dvipsnames]{xcolor}

\usepackage{float}
\usepackage{amsmath}

\newcommand{\modelname}{\emph{SwapAnything}}

\newcommand{\bs}[1]{{\boldsymbol{#1}}}

\newcommand{\eps}{\mathbf{\epsilon}}
\newcommand{\mask}{\mathbf{M_{src}}}

\newcommand{\selfoutput}{$\bs{\phi}$}
\newcommand{\selfmap}{$\bs{A}$}

\newcommand{\imagelatent}{$\bs{z}$}

\usepackage{hyperref}

\usepackage[capitalize]{cleveref}
\crefname{section}{Sec.}{Secs.}
\Crefname{section}{Section}{Sections}
\Crefname{table}{Table}{Tables}
\crefname{table}{Tab.}{Tabs.}

\usepackage{orcidlink}

\begin{document}

\title{
\modelname{}: Enabling Arbitrary Object Swapping in Personalized Image Editing
} 

\titlerunning{SwapAnything}

\author{
Jing Gu$^{1}$\thanks{This work was partly performed when the first author interned at Adobe.}\, Nanxuan Zhao$^{2}$\, Wei Xiong$^{2}$\, Qing Liu$^{2}$\, Zhifei Zhang$^{2}$ He Zhang$^{2}$\, Jianming Zhang$^{2}$\, HyunJoon Jung$^{2}$\, Yilin Wang$^{2}$\thanks{~Equal advising.}\, Xin Eric Wang$^{1\star\star}$\,
}

\authorrunning{J.~Gu, et al.}

\institute{\textsuperscript{1}University of California, Santa Cruz\; \textsuperscript{2}Adobe \\ 
\url{https://swap-anything.github.io/} }

\maketitle

\begin{figure}
    \centering
    \includegraphics[width=0.95\textwidth]{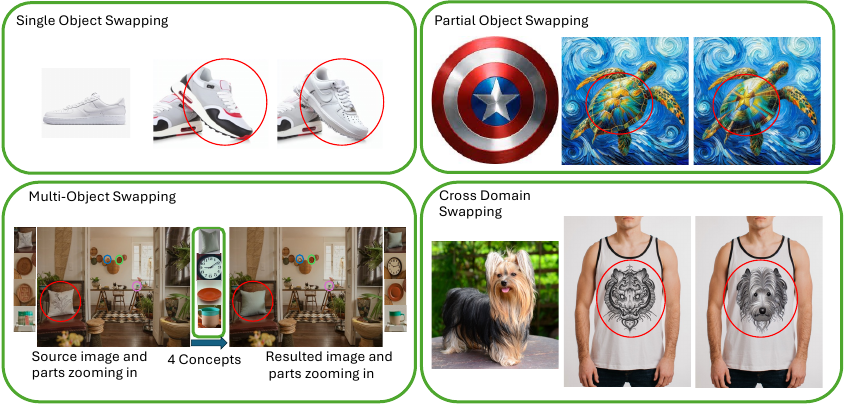}
    \caption{\textbf{\modelname{} results on various personalized image swapping tasks.} \modelname{} is adept at precise, arbitrary object replacement in a source image with a personalized reference, and achieves high-fidelity swapping results without influencing any context pixels. We demonstrate its general swapping effect in single-object, multi-object, partial-object, and cross-domain swapping tasks.
    }
    \label{fig:teaser} 
\end{figure}

\begin{abstract}
Effective editing of personal content holds a pivotal role in enabling individuals to express their creativity, weaving captivating narratives within their visual stories, and elevate the overall quality and impact of their visual content. 
Therefore, in this work, we introduce \modelname{}, a novel framework that can swap any objects in an image with personalized concepts given by the reference, while keeping the context unchanged.
Compared with existing methods for personalized subject swapping, \modelname{} has three unique advantages: (1) precise control of arbitrary objects and parts rather than the main subject, (2) more faithful preservation of context pixels, (3) better adaptation of the personalized concept to the image.
First, we propose \textit{targeted variable swapping} to apply region control over latent feature maps and swap masked variables for faithful context preservation and initial semantic concept swapping. Then, we introduce \textit{appearance adaptation}, to seamlessly adapt the semantic concept into the original image in terms of target location, shape, style, and content during the image generation process.
Extensive results on both human and automatic evaluation demonstrate significant improvements of our approach over baseline methods on personalized swapping. 
Furthermore, \modelname{} shows its precise and faithful swapping abilities across single object, multiple objects, partial object, and cross-domain swapping tasks. \modelname{} also achieves great performance on text-based swapping and tasks beyond swapping such as object insertion. 

\end{abstract}
    
\section{Introduction}
\label{sec:intro}

In today's digital age marked by the prolific creation and widespread sharing of personal content, generative models  \cite{rombach2022high,crowson2022vqgan,ding2021cogview,blattmann2022retrieval-diffusion} have risen as a potent tool for self-expression, storytelling, and amplifying the impact of visual narratives. Among existing image editing methods \cite{meng2021sdedit,hertz2022prompt,chen2023subject} for content creation that empowered individuals to convey their creative instincts, weave captivating narratives into their visual stories, and enhance the quality of their visual representations, personalize content swapping \cite{gu2023photoswap,li2023dreamedit,gu2023videoswap}, which targets at creating and compositing new images with user-specified visual concept, attracted significant interest due to its wide-ranging applications in E-commerce, entertainment, and professional editing. 

Achieving arbitrary personalized content swapping necessitates a deep understanding of the visual concept inherent to both the original and replacement subjects. There are several crucial challenges. Firstly, arbitrary swapping demands significantly greater flexibility from the swapping technique compared with swapping the main subject, due to the varied nature of the content being exchanged. Secondly, an ideal swap requires flawless preservation of the surrounding context pixels, ensuring that only the designated target area undergoes modification. The third challenge lies in accurately integrating the personalized content into the target image in a harmonious manner while preserving the original poses and styles. 

Existing works often fall short of addressing these challenges.  Most of existing research \cite{ruiz2023dream-booth,shi2023instantbooth,chen2023subject,jia2023taming,wang2024instantid}
are focused on personalized image synthesis, which seeks to create new images with personalized content. Although these approaches can synthesize high-fidelity content, they cannot edit or replace the visual content in an existing image. ~\cite{li2023dreamedit,cao2023masactrl,meng2021sdedit,avrahami2023blended} have tried to remove and regenerate the object via masks, they often struggle to adapt the new concept into the image. Recently studies~\cite{patashnik2023localizing,gu2023photoswap,tumanyan2022plug} focus on object swapping and replacement by 
tweaking intermediate variables affecting the object's features. However, this approach lacks the precision needed for localized object swapping, resulting in limited visual qualities. Besides, those methods mainly work on main subject swapping, and they are incapable of arbitrary object swapping.

To address these challenges, we introduce 
\modelname{}, a framework that utilizes pre-trained diffusion models to streamline personalized arbitrary object swapping. Unlike previous work, our work is designed for arbitrary swapping tasks with perfect context pixel preservation and harmonious object transition.
Our method begins by exploring an informative representation of the source image on a diffusion model. We found that various variables in the diffusion process especially latent features from U-net have a correspondent relation with the image. So we propose to keep the context pixels in the source image by preserving the correspondent part in those variables in the swapping process. This process is tailored to precisely swap specific areas, ensuring the preservation of other objects and the background's integrity. The object information in the source image is also selected for appearance adaptation. More specifically, location adaptation controls the location where the new concept should be swapped.
Style adaptation ensures stylistic harmony between the concept object and the original image, fostering a natural and cohesive visual presentation. Additionally, scale adaptation is introduced to modulate the target object’s shape, ensuring its congruence with the spatial and dimensional aspects of the source image. Last, content adaptation is crucial for smoothly generating the new concept, enabling a seamless blend that mitigates any artifacts or unnatural transitions. With these specialized adaptations, \modelname{} provides a heightened level of precision and refinement in the realm of object-driven image content swapping as shown in \cref{fig:teaser}. 

Our main contributions are: \textbf{1)} We propose \modelname{}, a general framework for both personalized swapping and text-based swapping on single object, multiple objects, partial object, and cross-domain object.
\textbf{2)} We identified key variables for content preservation and proposed targeted swapping for perfect background preservation. 
\textbf{3)} 
We designed a sophisticated appearance adaptation process to adapt the concept image into the source object.
\textbf{4)} Our approach has demonstrated exceptional performance through comprehensive qualitative assessments and quantitative analyses on swapping tasks and tasks beyond swapping such as insertion.

\section{Related Work}
\label{sec:related_works}

\subsection{Text-guided Image Editing}
With recent progress in diffusion model, text-guided Image editing has been largely explored~\cite{avrahami2023blended}.
Image editing driven by text aims to alter an existing image following the textual guidelines provided, while ensuring certain elements or features of the original image remain unchanged. Initial efforts using GAN models~\cite{karras2019style} were restricted to specific object domains. However, diffusion-oriented techniques~\cite{zhang2023adding,nichol2021glide,feng2023training-free} have surpassed this limitation, enabling text-driven image modifications. While these techniques produce captivating outcomes, many grapple with localized edits. As a result, manual masks~\cite{meng2021sdedit,zeng2022scenecomposer,meng2021sdedit} are often help to delineate the editing zones. Though employing cross-attention \cite{hertz2022prompt} or spatial attributes \cite{tumanyan2022plug} can achieve localized edits, they face challenges with flexible changes (like altering poses) and preserving the initial image composition. 

\subsection{Object Driven Image Editing}
In object-driven image generation, the object could be inverted into the texual space so to represented by a new token such as ``$*$'' ~\cite{gal2023ti, ruiz2023dream-booth, chen2023disenbooth, shi2023instantbooth, tewel2023keylocked}. 
Image editing guided by exemplars spans a vast array of uses. Much of the research in this area~\cite{wang2019example,huang2018multimodal,zhou2021cocosnet} falls under the umbrella of example-based image transformation tasks. These tasks leverage various sources of information, from stylized images~\cite{liu2021adaattn, deng2022stytr2, zhang2022inversion} and layouts~\cite{yang2022reco, li2023gligen, jahn2021high}, to skeletons~\cite{li2023gligen} and sketches or outlines~\cite{seo2022midms}. Given the versatility of stylized images, the art of image style adaptation~\cite{liao2017visual,zhang2020cross} has garnered significant interest. These methods primarily focus on establishing a detailed match between the input and reference visuals but falter when it comes to localized alterations. To facilitate localized changes, especially with flexible transformations, tools like bounding boxes and skeletons have been introduced. However, these tools often demand manual input, which can be challenging for users. A novel approach~\cite{yang2022paint} frames exemplar-driven image editing as a task of filling in gaps, maintaining the context while transferring semantic elements from the reference to the original image. While DreamEdit~\cite{li2023dreamedit} employs a step-by-step gap-filling method for object substitution, it doesn't effectively bridge the connection between the original and desired objects. In contrast, our method ensures that crucial features, like body movements and facial expressions, stay consistent. Meanwhile, Photoswap~\cite{gu2023photoswap}, CustomEdit~\cite{choi2023custom} and BLIP-Diffusion~\cite{li2023blip} achieved personalized object swapping. However, it does not keep the non-object pixel background intact. While our method directly focuses on the editing area without influencing other objects and background.

\section{Preliminary}

Diffusion Models belong to the family of generative models that are based on stochastic processes. They generate an image by iteratively reducing noise from an initial distribution.
Starting from a point of random noise denoted as $\bs{z}_T$, which follows a normal distribution $\bs{z}_T \sim \mathcal{N}(0, \mathbf{I})$, the diffusion model denoises each instance $\bs{z}_t$, thus producing $\bs{z}_{t-1}$.
Diffusion models predict and reverse the noise at each step in the diffusion sequence to arrive at the final denoised image. \par

In our research, we employ the pre-trained text-to-image diffusion framework Stable Diffusion~\cite{rombach2022high}. This model encodes images into a latent space and incrementally denoises the encoded latent representation. The Stable Diffusion model operates on a U-Net architecture~\cite{ronneberger2015u}, where the latent representation $\bs{z}_{t-1}$ at any given step is derived from the text prompt $P$ and the previous latent state $\bs{z}_t$, as indicated by the following equation:
\begin{equation}
\bs{z}_{t-1} = \bs{\epsilon\theta}(\bs{z}_t, P, t)
\label{eq:injection}
\end{equation}

This U-Net includes sequence of layers that repeatedly apply self-attention and cross-attention mechanisms. In self-attention, the latent image feature $z_t$, is first projected into query $Q_{self}$, $K_{self}$, $V_{self}$, which will be used to get self-attention map \selfmap{} and self-attention output \selfoutput{}.   
\begin{equation}
\begin{split}
    M & = \text{softmax} \left ( \frac{Q_{self}\cdot K_{self}^T}{\sqrt{d}} \right )  \\
    \phi & =  M \cdot V_{self}
\end{split}
\end{equation}

Meanwhile, for cross-attention layer, the feature out of previous self-attention layer is projected into $Q_{cross}$, while feature embedding of textual prompt is projected into $K_{cross}$ and $V_{cross}$.  
\begin{equation}
\begin{split}
    A & = \text{softmax} \left ( \frac{Q_{cross}\cdot K_{cross}^T}{\sqrt{d}} \right ) 
\end{split}
\end{equation}
where $A$ is the cross-attention map. In this work, we study the swapping of $A$, $M$, $\phi$ and $z$.

\begin{figure*}[t]
    \centering
    \includegraphics[width=1.0\textwidth]{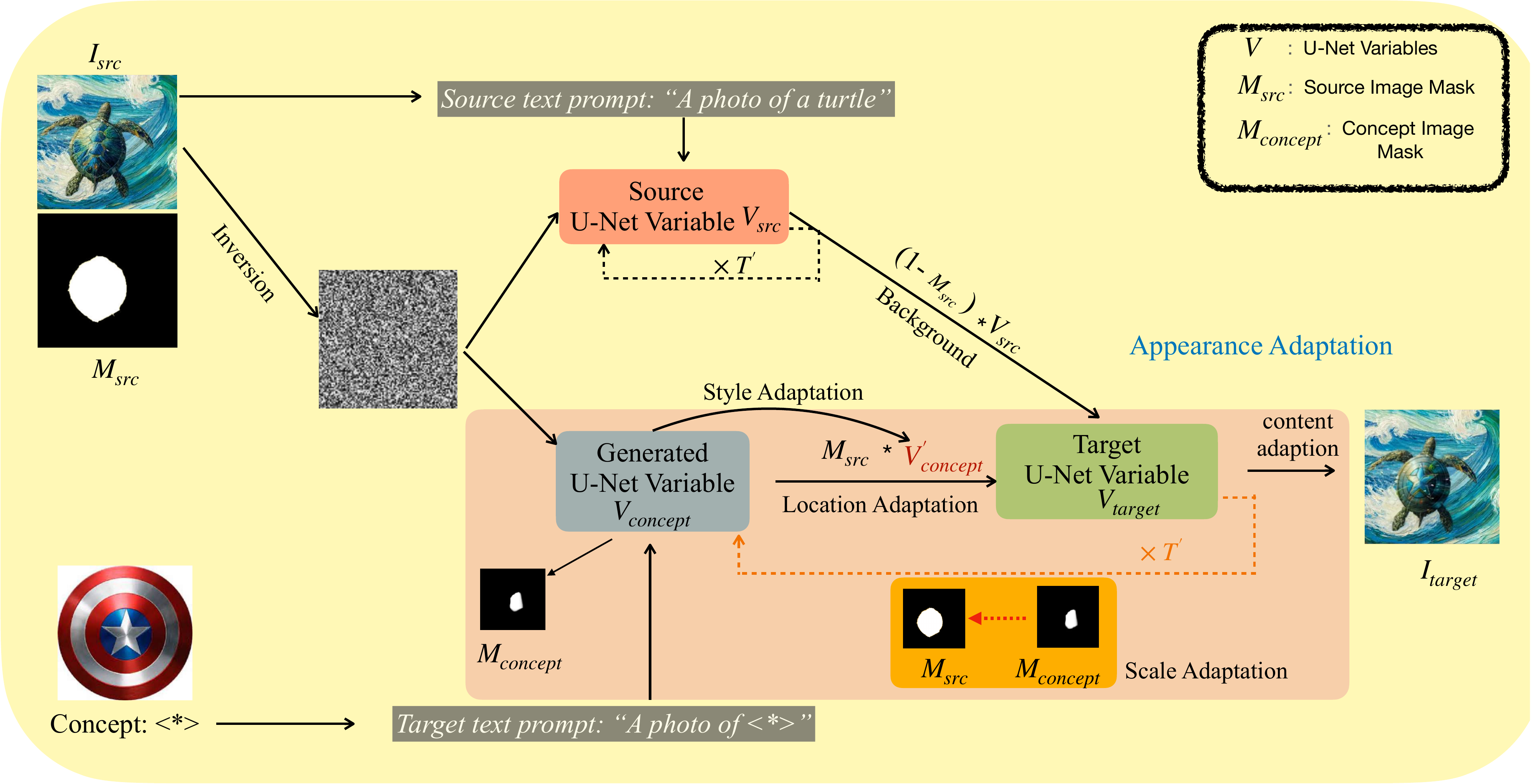}
    \caption{\textbf{Overview of \modelname{} on swapping a object from a source image ($I_{src}$) into a personalized concept ($<{*}>$) to get the target image ($I_{target}$).} The personalized concept is first converted into textual space to be treated as concept appearance.
    Meanwhile, the source image is first inverted into initial noise to obtain U-Net variables (including latent feature, attention map, and attention output). Targeted variable swapping preserves the context pixels in the source image. The appearance adaptation process then utilizes these informative variables to integrate the concept into the target image.}
    \label{fig:pipeline}
\end{figure*}

\section{\modelname{}}

\begin{figure}
    \centering
    \includegraphics[width=1.0\textwidth]{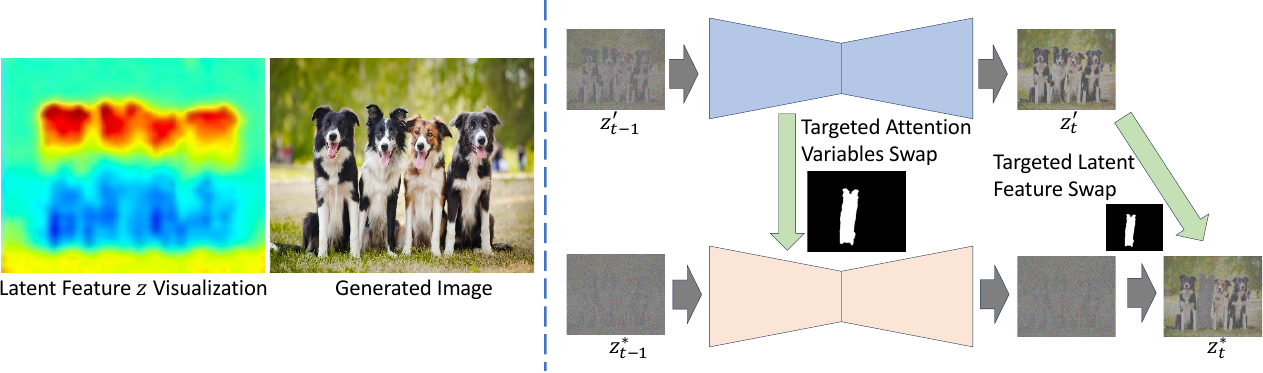}
    \caption{\textbf{Swapping process in \modelname{}.}
    The left part shows the correspondence between latent feature $z$ and the Generated image. The right part shows the procedure of targeted variable manipulation in the U-Net diffusion process.}
    \label{fig:swap-process}
\end{figure}

In this section, we introduce the \modelname{} framework that uses a diffusion model to swap on targeting area faithfully while keeping the context pixels unchanged. 
\cref{fig:pipeline} illustrates our framework's overall structure. For source image $I_{src}$, we first invert it to a latent noise and then obtain the feature representations $V_{src}$, which will be used during the target image $I_{src}$ generation process.
In \cref{subsec:targeted-variable-swapping}, we discuss how to preserve the non-target pixels in the source image perfectly, and how to select and transfer key information about the source image.
Following this, in \cref{subsec:appearance}, we introduce the appearance adaptation pipeline that uses the key information to integrate the new concept into the source image seamlessly.

\subsection{Targeted Variable Swapping}
\label{subsec:targeted-variable-swapping}

Intermediate variables in the U-Net of a diffusion model have been proven informative about the content of the generated image \cite{gu2023photoswap, hertz2022prompt, cao2023masactrl, tumanyan2022plug}. They usually focus on the study of variables inside of U-net structure such as attention map, and attention output, while the output of U-Net at each diffusion step, i.e., latent image feature $\bs{z}$ is not widely explored before. We argue that the latent image feature $\bs{z}$ contains more information on image content control. The image generation process for the latent diffusion model is achieved by denoising the $\bs{z}$ to arrive at a clear representation of a high-quality image, whereas all other variables inside of U-net indirectly affect the image by impacting $\bs{z}$. In contrast to simply swapping $\bs{z}$ like other variables, which would erase the new image's unique details and result in a mere duplication of the original image, our investigation revealed a significant correlation between the latent feature $\bs{z}$ and the produced image, including a pixel-level correspondence. In the left part of \cref{fig:swap-process}, we visualize the main component of the averaged latent feature $z$ across all diffusion steps. The finding that It has a \textbf{part-to-part} correspondence with the generated images indicates the potential of localized editing by manipulating the latent feature.

Consequently, we consider a strategy where only the context pixels within \imagelatent{} are altered, affecting solely the intended pixel. Although this method yields suboptimal outcomes, akin to merging two unrelated images, it prompts us to devise two remedial measures. Firstly, we suggest limiting the exchange of the latent feature to the initial stages of diffusion, allowing subsequent steps to smooth out any discordance in the latent space. Furthermore, our exploration into Unet's cross-attention map $\bs{M}$, self-attention map $\bs{A}$, and self-attention output $\bs{\psi}$ reveal their potential in mitigating artifacts. Swapping those facilitates the alignment of the latent features between the source and target images before the partial-swapping between them. In short, all the variables mentioned above in both the source image and target image generation process could be resized into the shape of the mask, where the mask can be utilized for the swapping process to get target variable:  
\begin{equation}
V_{target} = g(f(V_{src}) * (1-\mask{}) + f(V_{target}) * \mask{})
\label{eq:variableSwapping}
\end{equation}

Here $V$ includes latent feature $z$, and other assistant variable cross-attention map $\bs{M}$, self-attention map $\bs{A}$, and self-attention output $\bs{\phi}$. $f(\cdot)$ means the transformation process to the shape of the mask, while $g(\cdot)$ means the transformation back to the original space. For simplicity, we ignore all $f(\cdot)$ and $g(\cdot)$ in the following text. 
The content in the latent feature of the source image is changing as the diffusion process continues. Therefore, the location of the correspondent pixel in latent space may change over diffusion steps. A direct solution is to decode the latent feature $\bs{z}$ into an image at each step and extract the mask dynamically according to the object location in the generated image. However, we find that a changing mask usually confuses the model and leads to a less optimal performance. Therefore, we use the same high-quality mask through the diffusion process. We find that the mask could either be extracted from the source image directly using an off-the-shelf model or from the generation process as in~\cite{simsar2023lime, patashnik2023localizing}. Please check the Appendix for more details.

\subsection{Appearance Adaptation}
\label{subsec:appearance}

In this section, we introduce the appearance adaptation process that adapts the concept into the source image,
which necessitates meticulous adjustments across several dimensions: location, style, scale, and content. Our framework enhances realism and coherence in image manipulation, marking a significant advancement in the field.

\subsubsection{Location Adaptation}

Various intermediate variables have been proven to correlate with the final generated image. Although with great performance, it did not achieve local swap and thus the background is modified inevitably. As shown in \cref{fig:pipeline}, for each step, instead of directly swapping the whole variable, we conduct local swapping to only swapping the non-object position. Also, to enhance the swapping results, we further propose to conduct local swapping on the latent representation \imagelatent{} directly. $\mask{}$ is a 2-dimension variable containing 0 and 1. It is the same size as of the source image and value 1 marks the swapping location.
To simplify the expression, here we denote three U-Net variables attention map, attention output, and latent representation for the original image recovery process as $V_{src}$, denote the ones generated via target text prompt as $V_{concept}$, then we define the target variable used as $V_{target}^{bg}$ as the background information of the target variable, which can be obtained from \cref{eq:background}.
\begin{equation}
V_{target}^{bg} = V_{src} * (1-\mask{}) 
\label{eq:background}
\end{equation}
The non-masked area is the swapping target area, where the variable will be generated via the target text prompt to inject the concept appearance. Location adaptation extends beyond the object swapping tasks; we also discovered its profound capability for object insertion. For further details and results, please refer to the Appendix.

\subsubsection{Style Adaptation}
\label{subsec:style}

An ideal object swapping should keep the style unchanged. The object information in the generated variables is injected via the new concept token. Some style attributes could be already bound with the token. Therefore, solely generating the foreground information via the text prompt might lead to style inconsistency. 
Recently, \cite{karras2019style,park2019semantic} found that adding such normalization layers can help improve the conditional image generation quality because such activation functions modulation.  Unlike them, we employ the AdaIN (Adaptive Instance Normalization) to modulate the swapping features with spatial constraints. We follow \cref{eq:adain1} and \cref{eq:adain2} to denormalize the $V_{concept}$ with the mean and variance from $V_{src}$ in each time step for $V_{target}$ during the image generation process. As a result, we find that by modulating the concept feature, the generated content can adaptively follow the original style in the source image. 

\begin{equation}
    V_{concept}^{'} = MaskedAdaIN(V_{scr}, V_{concept}, M_{src})
    \label{eq:adain1}
\end{equation}
\begin{equation}
    V_{target}^{fg} = V_{concept}^{'}*M_{src}
\label{eq:adain2}
\end{equation}

MaskedAdaIN utilizes the mean and variance from the masked region in the AdaIN calculation. Then we have the blended feature representations for $V_{target}$:
\begin{equation}
    V_{target} = V_{target}^{fg} + V_{target}^  {bg}
\end{equation}

\subsubsection{Scale Adaptation}
\label{subsec:scale}

The proportion of an object compared to its environment and other elements in the image is crucial for coherence. A swapping result with improper scaling can disturb the aesthetic balance, resulting in a disjoint appearance of the image.

Guidance from an external classifier in the inference process of Diffusion models could influence the diffusion noise to control the generated image.
\cite{epstein2023selfguidance} shows that the guidance can also be used on the attention map to control the generation. Similarly, we adapt the mask guidance \cref{eq:scale} to better align the shape between the source object and the target object. 
\begin{align}
    \hat{\epsilon}_t &= (1 + s) \epsilon_\theta(z_t; t, y) - s \epsilon_\theta(z_t; t, \emptyset) \nonumber \\
    &\quad + v \sigma_t \nabla_{z_t} \|M_{src}- Shape(M_{src})\left( k\right)\|_1
    \label{eq:scale}
\end{align}
where $s$ is the classifier-free guidance strength and $v$ is an additional guidance weight for $g$.  As with classifier guidance, we scale by $\sigma_t$ to convert the score function to a prediction of $\eps_t$. 
$Shape(\mask{})\left( k\right)$ denotes the object shape as identified in the cross-attention layer.
Here the energy function $g$ is set as $\|\mask{} - Shape(\mask{})\left( k\right)\|_1$ to calculate the shape difference between the original object mask and the extracted shape of object token $k$ in the attention layer, which indicates the deviation between the ideal shape and shape during the diffusion process.

\subsubsection{Content Adaptation}
\label{subsec:content}
A binary mask without smoothing has a high-frequency transition at the edge — it jumps abruptly from 0 to 1. When used to merge two intermediate variables from two different diffusion processes, this can result in high-frequency artifacts at the boundary, such as jagged edges or a halo effect. Smoothing the mask transitions these high frequencies into lower frequencies, which blends the images more naturally and eliminates such artifacts. A smooth mask creates a feathering effect at the edges of the transition. This makes the merged area appear more coherent as if the two images naturally blend into each other rather than being cut off abruptly. Therefore, for the diffusion process, we specifically design two masks according to the feature of diffusion models. 

Without this smoothing, the boundary between the images would be sharply defined, leading to a jarring and unnatural appearance. The Gaussian Blur softens the edges, blending the images more seamlessly.
To augment this improvement, we introduce two smoothing techniques for binary masks, applied across both spatial dimensions and temporal steps. These techniques serve to refine the swapping process, mitigating artifacts and ensuring a smoother, more natural integration of the swapped regions. This results in an enriched visual output, seamlessly blending the inserted objects or object parts into the overall image composition.

\noindent\textbf{Linear Boundary Interpolation}: This is a process where the sharp transition between the area with 1s and the area with 0s in your binary array is made gradual. One way to achieve this is by using a convolution with a smoothing kernel (like a Gaussian kernel) that will average the values in the vicinity of each point, effectively creating a gradient at the boundary.
\begin{align}
    \delta(\mask{}) &= \mask{} \oplus K \nonumber\\
    S &= \delta(\mask{}) \ast G \nonumber\\
    S'[i, j] &= 
    \begin{cases} 
        1 & \text{if } M[i, j] = 1 \\
        S[i, j] & \text{otherwise}
    \end{cases}
\end{align}

The dilation of the mask $\mask{}$ using the structuring element K, where denotes the dilation operation and G is the Gaussian kernel. The asterisk $*$ denotes the convolution operation.
$S'$ is the final soft mask.

\noindent\textbf{Gradual Boundary Transition}: This involves generating a sequence of arrays where the value of 1 does not appear immediately but increases incrementally from 0 to 1. This can be achieved by interpolating between 0 and 1 across the sequence of arrays.
\begin{equation}
\mask{}(x, y) = 
\begin{cases}
\mask{}(x, y) \cdot \frac{t}{K}, & \text{if } t \leq K \text{ and } \mask{}(x, y) = 1 \\
\mask{}(x, y), & \text{otherwise}
\end{cases}
\end{equation}
In this equation, the value of \( \mask{}(x, y) \) is assumed to be 1 in the center area and 0 elsewhere. For the central region, the value linearly increases from 0 to 1 over the first \( K \) steps. For the rest of the mask, the original value \( \mask{}(x, y) \) remains unchanged.

Several prevalent backbone diffusion models, including Stable Diffusion 2.1, are restricted to processing images in a square format. Resizing images to fit a square dimension can lead to substantial content distortion, adversely affecting the editing outcomes. Nevertheless, our findings demonstrate that our method exhibits a remarkable capacity for adaptation, allowing it to \textbf{process images of any aspect ratio} without compromise. As documented in this paper, we present all images in various ratios.

\section{Evaluation}
\label{sec:evaluation}

\begin{figure*}[ht]
    \centering
    \includegraphics[width=1.0\textwidth]{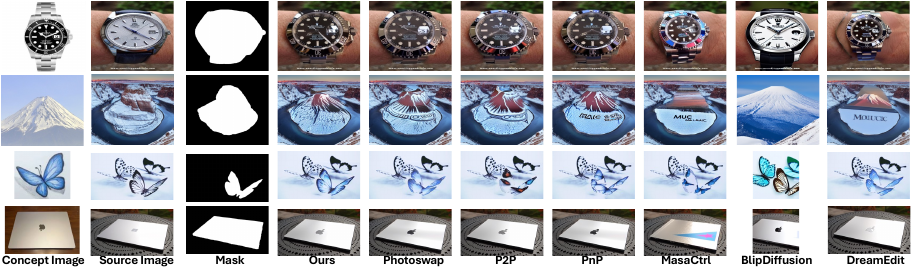}
    \caption{\textbf{Qualitative comparison with different baselines.} Note that those baseline methods were already equipped with some components of \modelname{} for precise control of the swapping region. Please check \cref{subsec:implementation-details} for details.}
    \label{fig:single-subject-result}
\end{figure*}

\subsection{Implementation Details}
\label{subsec:implementation-details}

Here we introduce implementation details.
In our paper, we used Stable Diffusion 2.1 as the pre-trained text-to-image diffusion model. DreamBooth~\cite{ruiz2023dream-booth} is used to convert the concept into textual space. 
We used null-text inversion~\cite{mokady2022null} based on DDIM inversion~\cite{song2020denoising} to boost the accuracy.
There is \textbf{no additional operation} for single-object, partial object, cross domain swapping. Multi-object swapping is achieved by conducting swapping operation on the previous swapped image. Please check Appendix for evaluation dataset details.

\noindent\textbf{Baselines setting}. Photoswap\cite{gu2023photoswap}, P2P\cite{hertz2022prompt}, PnP\cite{tumanyan2022plug}, MasaCtrl\cite{cao2023masactrl} are attention variable based image editing methods, which are also compatible with our proposed Masked Latent Blending in \cref{subsec:targeted-variable-swapping} and Location Adaptation in \cref{subsec:appearance}. Therefore we boost their performance with those additional components. Otherwise, their performance would be much worse. Please see appendix for comparisons with their original methods, and the implementation details. Also, please check appendix for its performance on object insertion.

\begin{table*}
\centering
    \setlength{\tabcolsep}{2.0pt}
    \caption{\textbf{Human evaluation results.} We show the human preference between results generated by our method and the baseline methods. SS means Object Swapping, BP means Background Preservation, and OQ means Overall Quality. SG means Object Gesture. For the baseline methods, PS means Photoswap; MC means MasaCtrl; BP means BlipDiffusion; DE means DreamEdit; CP means CopyPaste.
    }
    \resizebox{\linewidth}{!}{
    \begin{tabular}{c|ccc|ccc|ccc|ccc|ccc|ccc|ccc}
        \toprule
        ~ & Ours & PS & Tie & Ours & P2P & Tie & Ours & PnP & Tie  &
        Ours & MC & Tie &
        Ours & BP & Tie & Ours& DE& Tie & Ours & CP & Tie\\
        \midrule
        SS ~ & \textbf{52.3} & 12.5 & 35.2 & \textbf{55.3} & 17.9 & 26.8 & \textbf{52.3} & 29.9 & 11.5 & \textbf{64.3} & 20.0 & 15.7 & \textbf{55.3} & 16.5 & 28.2 & \textbf{55.1}& 20.1&24.8& \textbf{60.1}& 17.3&22.6\\
        SG ~ & \textbf{44.5} & 34.0 & 21.5 & \textbf{49.5} & 32.0 & 18.5 & \textbf{55.5} & 33.0 & 11.5 & \textbf{65.3} & 18.8 & 15.9 & \textbf{41.5} & 39.6 & 18.9 &\textbf{44.3} &18.8&36.9&\textbf{54.3}&20.5&25.2\\
        BP ~ & \textbf{41.5} & 32.0 & 26.5 & \textbf{44.5} & 28.9 & 26.6 & \textbf{50.2} & 20.9 & 28.9 & \textbf{55.2} & 23.1 & 21.7 & \textbf{40.0} & 27.5 & 32.5 & \textbf{44.0} &18.9&37.1&\textbf{54.1}&13.1& 32.8\\
        OQ ~ & \textbf{49.3} & 27.8 & 22.9 & \textbf{55.0} & 27.2 & 17.8 & \textbf{52.5} & 22.9 & 24.6 & \textbf{59.4} & 20.1 & 20.5 & \textbf{48.1} & 25.3 & 26.6 &\textbf{53.1} &19.9&27.0&\textbf{71.1}&10.5& 18.4\\
        \bottomrule
    \end{tabular}
    }
    \label{table:human-evaluation}
\end{table*}

\subsection{Single-object Swapping}

We consider human evaluation to be the main quantitative performance measurement. A successful swap should keep the non-object area unchanged, change the object identity to target, and keep the gesture the same as the source object. As in \cref{table:human-evaluation}, our model consistently outperforms baselines across all metrics. \cref{fig:single-subject-result} shows the qualitative comparison for both human and non-human images. Thanks to the addition of our targeting variable swapping and location adaptation, all attention-manipulation based baselines also achieved perfect background preservation and some level of localized swapping result. \modelname{} yield a much better appearance adaptation result.

\subsection{Multi-object Swapping}

\begin{figure*}
    \centering
    \includegraphics[width=\textwidth]{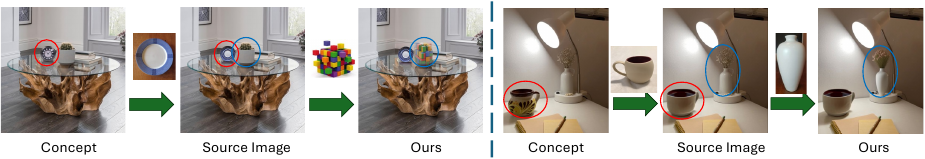}
    \caption{\textbf{Multi-object swapping results of \modelname{}.} Our method could easily swap multiple objects via swapping one object at a time. Note that the red circle means the target object to be replaced. The same color means a pair of concept and target for object swapping.}
    \label{fig:multi-objects}
\end{figure*}

As is shown in \cref{fig:multi-objects}, multi-object swapping is simply achieved via repeating single-object swapping, which further highlights its versatility and efficiency. Multi-object swapping is a natural outcome of our targeted variable swapping, whereas previous methods struggle to achieve satisfactory results. Without perfect context pixel preservation, the unwanted image modification would accumulate as the swapping continues. Please check the appendix for more results and comparison with baselines.

\subsection{Partial Object Swapping}
\begin{figure}
    \centering
    \includegraphics[width=1.0\textwidth]{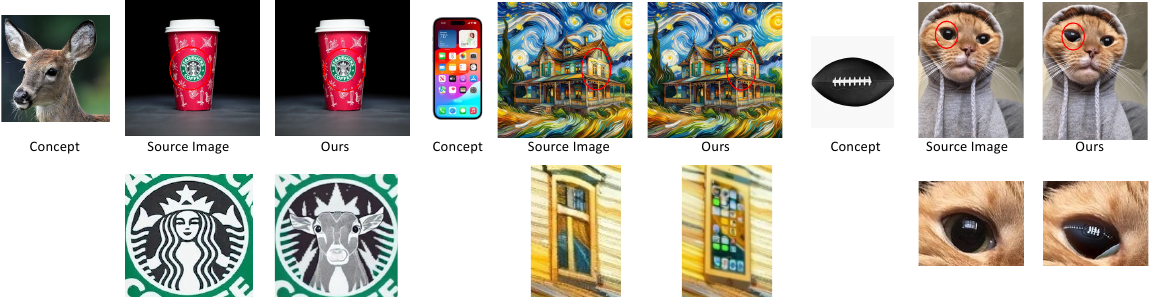}
    \caption{ \textbf{Results on partial object swapping.} \modelname{} can swap partial object that is tightly connected with other parts and adapt seamlessly to the source image. The second row is the zoom-in images of the swapping part in the first row.
    }
    \label{fig:subject-part}
\end{figure}

As is shown in \cref{fig:subject-part}, \modelname{} achieved a great performance on swap a part of a whole object, even when the targeting area is very small. Meanwhile, all other baselines failed to achieve such results. Please check appendix for more results.

\subsection{Cross-domain Swapping}
\begin{figure}
    \centering
    \includegraphics[width=1.0\textwidth]{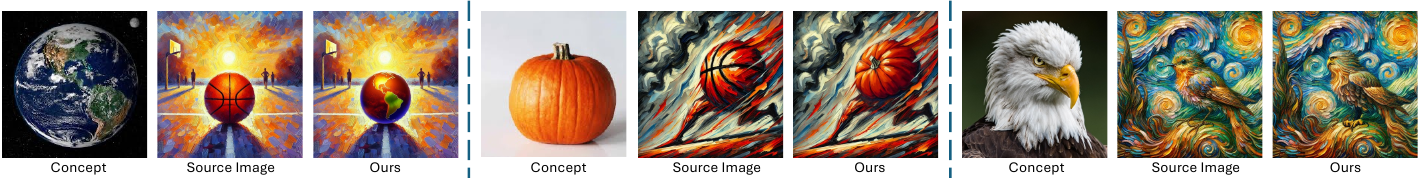}
    \caption{ \textbf{Results on cross-domain object swapping.} With a variety of source images, including graphic, black-white photos, and oil paintings, the framework seamlessly integrates concept objects taken from regular images into these diverse source images.
    }
    \label{fig:cross-domain}
\end{figure}

\cref{fig:cross-domain} demonstrates that \modelname{} can adeptly handle a range of stylized source images, successfully adapting concept objects to match the desired style within the source image while seamlessly transferring identity into the generated images. For instance, when the source image is a photo of a certain style, yet \modelname{} skillfully generates the same painting style featuring personalities like ``Charles Darwin'' and ``J. Robert Oppenheimer''. Notably, all the concept images are regular, unstyled photos, underscoring the model's ability to blend different styles and identities effectively.

\subsection{Text-based Swapping}
\begin{figure}[h]
    \centering
    \includegraphics[width=0.95\textwidth]{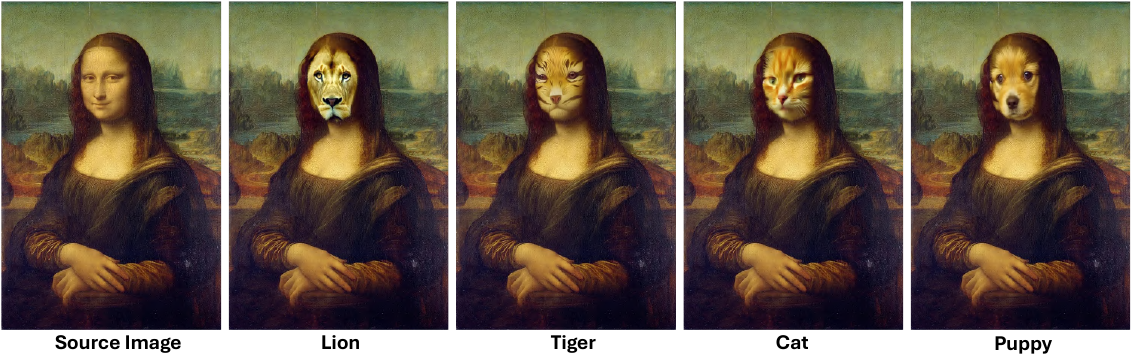}
    \caption{\textbf{Results on text-based swapping.} Besides swapping from concept image, \modelname{} is also capable of swapping an object described in text into the image.}
    \label{fig:text-based}
\end{figure}

As shown in Fig.~\ref{fig:text-based}, besides personalized swapping, our method can also do text-based swapping, swapping an object in the source image with another described in text. This can be achieved by simply replacing the personalized concept token $*$ with a text prompt, e.g., ``A photo of $new\_obj$''. 

\subsection{Ablation Study}

\begin{figure}
    \centering
    \includegraphics[width=1.0\textwidth]{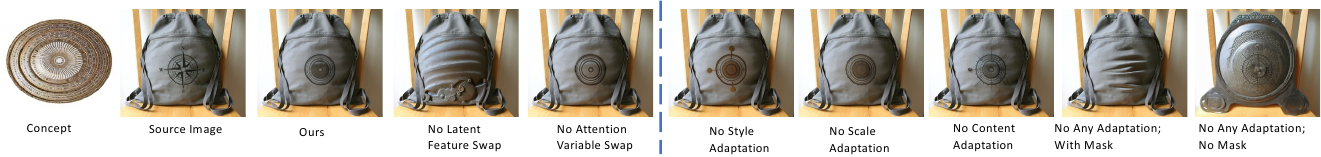}
    \caption{\textbf{Ablation study.} The left part shows the effect of swapping, and the right part shows the effect of adaptation and mask.}
    \label{fig:ablation}
\end{figure}

\cref{fig:ablation} show the effect of the components in \modelname{}. From the left part, we see that without latent feature swap, even with a mask and attention variable swap, the context pixel is still changed. Both latent feature and attention variable has effect of information preservation when compared with the result of no swap. The effect of adaptation and mask is presented on the right part. With style adaptation, the visual texture is closer to the source image. Without scale adaptation, the shape is not well aligned and artifacts appear in the neck part. When without any adaptation, the generated image is much less connected the source image regarding the swapping area. When without mask, both background and targeting area are changed, which leads to a different image. 

\subsection{Ethics Discussion}

While \modelname{} achieves impressive performance across various visual tasks, it also raises potential ethical concerns. Like other advanced image editing technologies, \modelname{} could be misused to create deceptive content. To mitigate such risks, implementing technologies like digital watermarking can help track modifications and authenticate the sources of images, assisting users in distinguishing between genuine and altered content. Moreover, advocating for clear guidelines and regulations that govern the ethical use of image manipulation technologies is essential. These measures ensure responsible development and use, fostering a balance between innovation and ethical considerations. Here, we emphasize the importance of applications involving human face swapping. We strongly discourage using our methods on human faces without consent, and we are committed to preventing misinformation and harm to any individual or community. We enforce strict consent protocols in related tasks, requiring explicit permission from individuals before their images are used, especially in sensitive applications like face-swapping.

\section{Conclusion}
\modelname{} represents a notable breakthrough in the realm of object swapping. Swapping latent features and attention variables in the diffusion model ensures the retention of crucial information within the generated image. Through a targeted manipulation, we achieved a perfect background preservation. Additionally, we have introduced a sophisticated appearance adaptation process designed to seamlessly integrate the concept into the context of the source image. Consequently, \modelname{} is equipped to handle a diverse array of object swapping challenges. In the future, we plan to extend our framework to 3D/video personalized object swapping tasks.

\bibliographystyle{splncs04}
\bibliography{main}

\appendix

\clearpage

\section{Object Insertion}
\begin{figure}[hbt]
    \centering
    \includegraphics[width=0.95\textwidth]{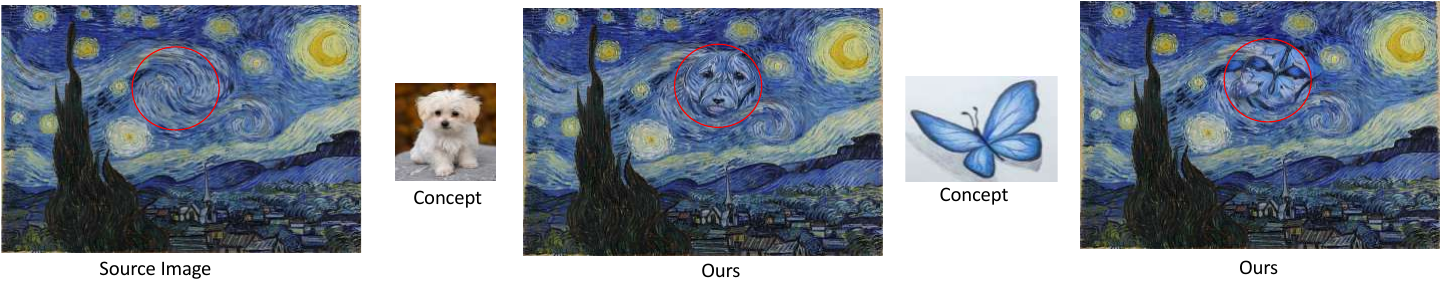}
    \caption{\textbf{Results on object insertion.} \modelname{} can insert and adapt an object into a certain location of an image.}
    \label{fig:insertion}
\end{figure}

\modelname{} is a general framework and is also capable of object insertion. With the same process as single-object swapping, we could insert and adapt a concept into background pixels, while preserving the composition and style of the source image. In~\cref{fig:insertion}, we insert a puppy and a butterfly into The Starry Night from Vincent van Gogh.

\section{Variable Swapping Details}
We use Stable Diffusion 2.1 as the pre-trained text-to-image diffusion model. DreamBooth~\cite{ruiz2023dream-booth} is used to convert the concept into textual space. 
The learning rate for this process is set at 1e-6, and we use the Adawm optimizer for 800 steps. The U-net and the text encoder are fine-tuned during this process, typically taking about 2 minutes on a machine equipped with 8 A100 GPUs. The target prompt is essentially the source prompt with a swap in object tokens to introduce a new concept.

For object mask, we first detect the object with Grounding DINO~\cite{liu2023grounding} and then extract the mask using Segment Anything~\cite{Kirillov_2023_ICCV}. 
For the targeting variable swapping, we do 30 for latent image feature $z$, 20 steps for cross-attention map, 25 for the self-attention map, and 10 for the self-attention output, we conduct swapping in all U-Net layer.

For area mask smooth, we first enlarge the masked areas using a dilation operation with an elliptical kernel, which can be adjusted in size. After dilation, the mask edges are smoothed using a Gaussian blur, creating a gradient effect at the boundaries. For the smooth over diffusion step, we linearly increase the mask rate from 0 to 1 during the first 30 steps. For a better understanding, we mark the masked area using a circle in most figures in this paper.

\section{Adaptation Details}
\textbf{Style Adaptation.}
This operation adjusts the mean and variance of content image features to match those of the style features, facilitating the transfer of artistic styles onto content images. The AdaIN technique is renowned for its efficiency and flexibility, making it a go-to choice for real-time style transfer and artistic image manipulation. Building on this, we introduce Masked-AdaIN. Unlike traditional AdaIN which applies style alignment across the entire image, Masked-AdaIN focuses this alignment only on a specific target area. In this approach, mean and variance calculations are exclusively performed on the designated masked area, allowing for more precise and localized style transfers.

\noindent\textbf{Scale Adaptation.} 
We adapt the scale of the object in latent space to the shape of the mask. The object shape is indicated in the cross-attention map at each diffusion step~\cite{gu2023photoswap, hertz2022prompt}. $Shape(\mask{})(k)$ means the attention map for object text token $k$, which is obtained through binary-like transformation to the attention map. We apply a threshold of 0.4 after using sigmoid to normalize the attention value between 0 and 1. 

\noindent\textbf{Content Adaptation.} 
In the Linear Boundary Interpolation process, the structuring element $K$ is a predefined shape used in the dilation process to create the dilated image. The structuring element $K$ slides over the binary mask $\mask{}$ and at each position. If at least one pixel under $K$ is 1, the pixel in the output image under the center of $K$ is set to 1. This operation typically results in the enlargement of the regions with 1s in the binary mask, effectively smoothing the boundary and filling small holes and gaps. The subsequent convolution with a Gaussian kernel $G$ further smooths the mask by averaging values in the vicinity of each point, thereby creating a gradient effect. The combination of dilation and Gaussian smoothing prepares the mask $S'$ for linear boundary interpolation, where the sharp transitions are made gradual, and the final soft mask 
$S'$ is obtained by selectively setting pixels to 1 based on the original mask and the smoothed values. 
In Gradual Boundary Transition, we set the transition step parameter as 30 to anneal $\mask{}$ from 0 to the set value.

\section{Dataset}
We conducted experiments on both human and non-human objects. For human swapping, we collect 50 faces as concepts. We also collected 500 images containing 1 or more people as the source images. For non-human object, we include DreamEdit~\cite{li2023dreamedit} dataset and more concepts and its corresponding source images. In total, we aggregated 1,000 images.

\section{Human Evaluation Interface}
\label{human-eval-interface}

\begin{figure*}[hbt]
    \centering
    \includegraphics[width=0.9\textwidth]{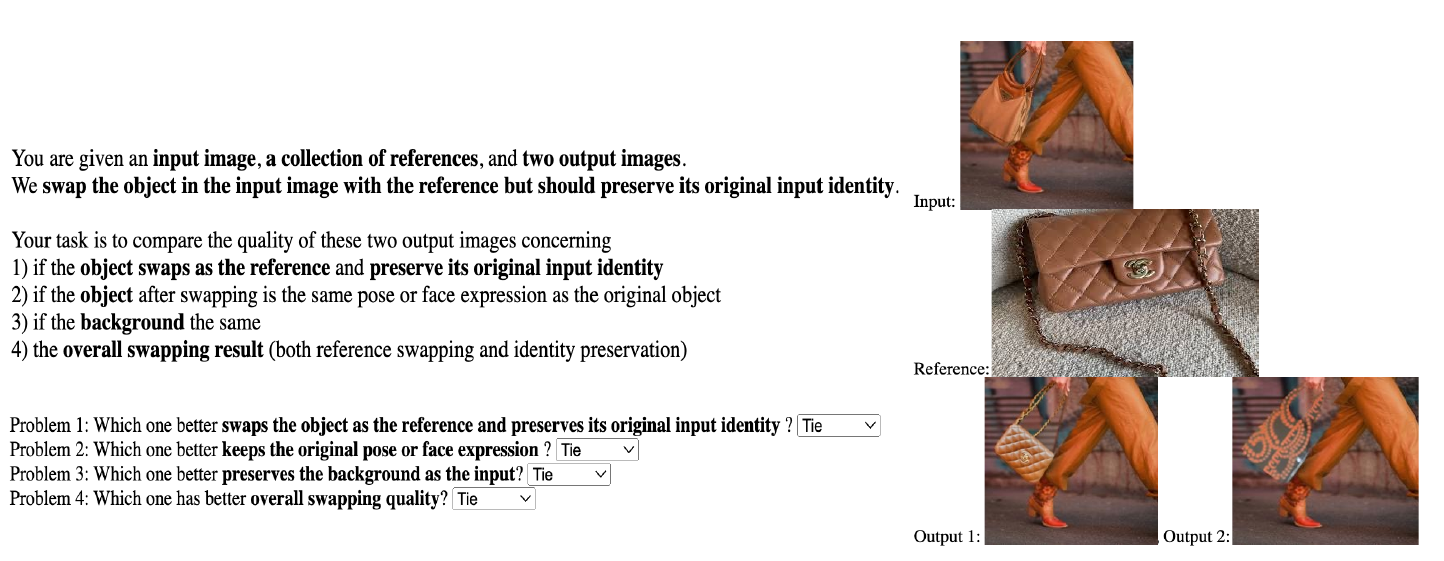}
    \caption{\textbf{The illustration of the user study interface.}}
    \label{fig:human-study-interface}
\end{figure*}

Amazon Turker was presented with one reference image mainly containing the concept subject, one source image to be swapped, and two generated images from \modelname{} and a baseline.

\section{More Qualitative Results}

\begin{figure*}
    \centering
    \includegraphics[width=0.95\textwidth]{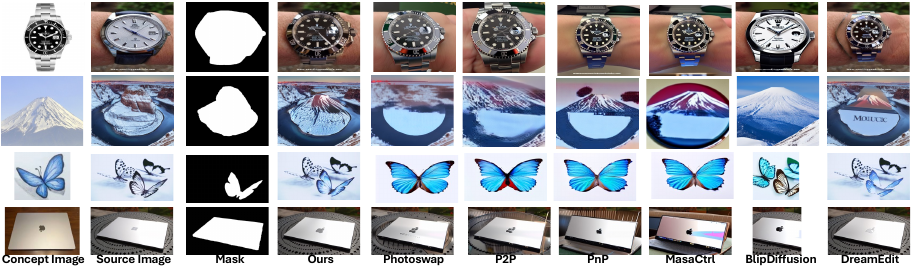}
    \caption{\textbf{Comparison on single-object swapping with baselines in their original components.} Please zoom in for a clear visual result. }
    \label{fig:more-single}
\end{figure*}
Here we first show the comparison with baselines in their original setting on single-object swapping. On other more challenging tasks,
we also show the results of Photoswap since it is the state-of-the-art method of subject swapping. For the implementation details, we use external mask to help the inpainting process in DreamEdit. \modelname{} is also compared with BlipDiffusion~\cite{li2023blip}. Photoswap, P2P, PnP, and MasaCtrl, DreamEdit were equipped with the same DreamBooth model to grasp the new concept. Note that this would also indirectly include comparison with CustomEdit~\cite{choi2023custom}, since it also achieved personalized object swapping via equiping P2P with concept learning. CopyPaste involves directly transplanting the concept object in the concept image into the source object's position.

\noindent\textbf{Single-object Swapping.} \cref{fig:more-single} shows comparisons between \modelname{} and baselines. \modelname{} consistently outperforms other models in terms of background preservation, identity swapping, and overall quality. Note that there is also a huge performance gap between some baselines and their counterpart in Fig. 4 in the main paper, which further validates the efficacy of targeted variable swapping and location adaptation, which was applied to Photoswap, P2P, PnP, and MasaCtrl in Fig. 4 in the main paper.

\begin{figure}
    \centering
    \includegraphics[width=0.95\textwidth]{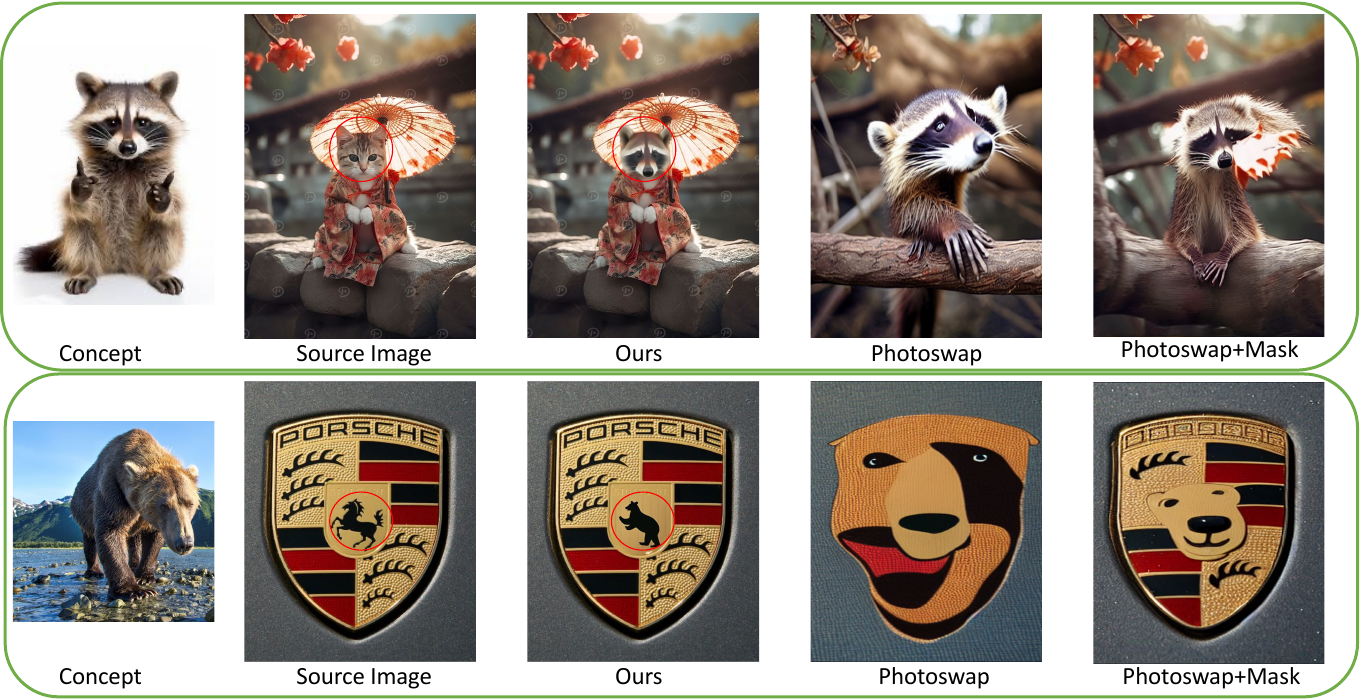}
    \caption{\textbf{Comparison with Photoswap on partial object swapping and cross-domain swapping.} The upper part shows \modelname{} could localize the swapping area while Photoswap inevitably modified the background. In the lower part, \modelname{} adapts a bear into the style of a logo, while Photoswap failed on this cross-domain swapping task. }
    \label{fig:more-cross-and-partial}
\end{figure}

\noindent\textbf{Partial Object Swapping.} As in \cref{fig:more-cross-and-partial}, our method precisely swaps the cat head with a raccoon head harmoniously without influencing other pixels. Meanwhile, Photoswap swaps the whole body and modified the context pixels. When our proposed masked variable swapping is added, Photoswap achieves a better background preservation performance.

\noindent\textbf{Cross-domain Swapping.} \modelname{} is capable of swapping between styles and textual. In \cref{fig:more-cross-and-partial}, a bear is adapted into a logo while keeping the gesture of the source object horse. Meanwhile, Photoswap fails to complete the challenging task. Also, when masked variable swapping is added, Photoswap achieves a better adaptation performance.

\noindent\textbf{Multi-object Swapping.} Multi-object swapping is a big step after single-object swapping.  
First, previous methods usually have a background modification such that continuous editing would accumulate unwanted distortion, which leads to a totally different image and fails the task of swapping. The second issue is that previous methods are usually designed for main subject swapping and do not pay attention to other objects. In this case, the objects in the following swapping steps could disappear in the previous swapping process.

\section{More Quantitative Results}

\begin{table}
\centering
\small
    \caption{\textbf{Automatic evaluation results.} \modelname{} outperforms all other methods across all metrics.
    }
    \resizebox{0.9\linewidth}{!}{
    \begin{tabular}{c|cccccccc}
        \toprule
        ~ & Ours& Photoswap& P2P & PnP & MasaCtrl & BlipDiffusion & DreamEdit & CopyPaste\\
        \midrule
        $DINO_{fore}$ ~ & \textbf{0.61}&0.55&0.47&0.49&0.29& 0.44 & 0.52 & 0.56\\
        $CLIP_{fore}$ ~ & \textbf{0.79}&0.53&0.71&0.73&0.46& 0.54& 0.61 & 0.75\\
        $DINO_{back}$ ~ & \textbf{0.79}&0.75&0.68&0.64&0.71&0.71 & 0.76 & 0.77\\
        $CLIP_{back}$ ~ & \textbf{0.89}&0.86&0.75&0.70&0.67& 0.76& 0.82 &0.79\\
        \bottomrule
    \end{tabular}
    }
    \label{table:automatic-evaluation}
\end{table}

We also conducted automatic evaluation. Following \cite{ruiz2023dream-booth, li2023dreamedit, gu2023photoswap}, we employ both DINO and CLIP-I as tools to evaluate the quality of the images generated. These two metrics serve as complementary indicators to the results obtained from human evaluations. As in~\cref{table:automatic-evaluation}, \modelname{} outperforms all other baselines in terms of both subject identity swapping and background preservation, which is consistent with the results of human evaluation.

\cref{table:more-human-evaluation} shows results on human evaluation on both human and non-human images on top baselines. PS means Photoswap~\cite{gu2023photoswap}; P2P means Prompt-to-Prompt~\cite{hertz2022prompt}; PnP means Plug-and-Play~\cite{tumanyan2022plug}; DE means DreamEdit~\cite{li2023dreamedit}. We also conduct comparisons with another baseline PbE, Paint-by-Example~\cite{yang2022paint}.

\begin{table}
\centering
\small
    \setlength{\tabcolsep}{2.0pt}
    \caption{\textbf{User study results.} The 2nd to 5th rows and 6th to 9th rows show results on human objects and non-human objects.
    }
    \begin{tabular}{c|ccc|ccc|ccc|ccc|ccc}
        \toprule
        ~ & Ours & PS & Tie & Ours & P2P & Tie & Ours & PnP & Tie & Ours& DE& Tie & Ours& PbE&Tie\\
        \midrule
        SS ~ & \textbf{59.0} & 10.0&  31.0  & \textbf{52.7} & 18.2 & 24.1 & \textbf{58.8} & 29.2 & 12.0 & \textbf{53.4}& 16.5&30.1 &\textbf{62.1}&12.0&28.0\\
        SG ~ & \textbf{44.0} & 33.7 & 22.3 & \textbf{54.5} &29.1  & 16.4 & \textbf{61.6} &33.3 &5.1  &\textbf{42.4} &17.2&40.4&\textbf{73.1}&15.8&12.0\\
        BP ~ & \textbf{45.4} &32.2 & 22.4 & \textbf{49.9} &26.9 & 23.2& \textbf{49.7} & 22.0 & 28.3 &  \textbf{43.8} & 18.0& 38.2&\textbf{42.1}& 31.3& 27.0\\
        OQ ~ & \textbf{47.3} &24.3 &28.4 & \textbf{58.4} &23.3 & 18.3& \textbf{51.6} & 31.1 & 17.3 &\textbf{47.5} &26.5& 26.0&\textbf{52.1}&21.9&30.0\\
        \midrule
        SS ~ & \textbf{45.6} & 15.0 & 39.4 & \textbf{52.9} & 17.6  & 29.5 & \textbf{45.8} & 30.6 & 23.6 & \textbf{56.8}&23.7 &19.5&\textbf{58.1}&12.0&27.8\\
        SG ~ & \textbf{45.0} & 34.3 & 20.7 & \textbf{44.5} & 34.9 & 20.6 & \textbf{49.4} &32.7 & 17.9 &\textbf{46.2} &20.4& 33.4&\textbf{69.3}&15.0&14.8\\
        BP ~ & \textbf{37.6} & 31.8 & 30.6 & \textbf{39.1} & 30.9 & 30.0 & \textbf{50.7} & 19.8& 29.5 &  \textbf{44.2} &19.8& 36.0&\textbf{40.5}&31.5&27.6\\
        OQ ~ & \textbf{51.3} & 31.3& 17.4& \textbf{51.6} &31.1 &17.3 & \textbf{47.5} & 26.5& 26.0 &\textbf{48.1} &21.2& 30.7&\textbf{48.1}&18.3&29.6\\
        \bottomrule
    \end{tabular}
    \label{table:more-human-evaluation}
\end{table}

\section{Failure Cases}

\begin{figure}[t]
    \centering
    \includegraphics[width=\textwidth]{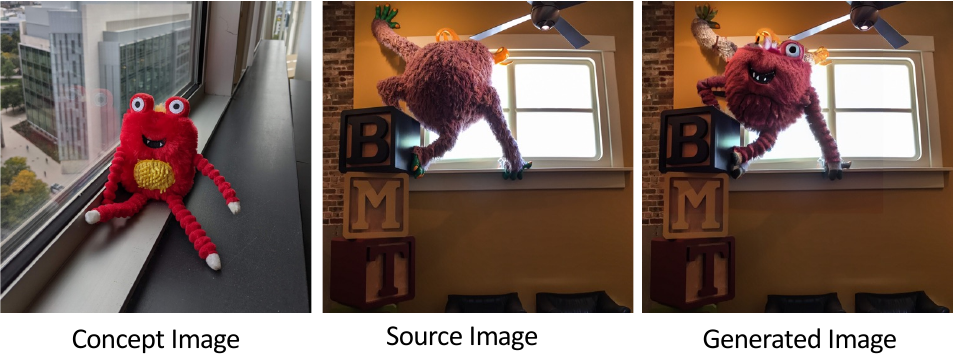}
    \caption{\textbf{Examples of failure cases.} The model sometimes struggles to keep the details inside the mask area and could fail if the object has a high degree of freedom.}
    \label{fig:failure-cases}
\end{figure}

We highlight one common failure scenario encountered in our experiments. The challenge arises when dealing with subjects that exhibit a high degree of variability or freedom of movement. In such cases, as shown in~\cref{fig:failure-cases}, accurately replicating the concept subject becomes difficult. To address this, we are considering the implementation of explicit alignment, which we aim to explore in our future work.

\end{document}